\documentclass[runningheads]{llncs}

\usepackage[T1]{fontenc}
\usepackage{graphicx}
\usepackage{bbm}
\usepackage{multirow}
\usepackage{makecell}
\usepackage{booktabs}
\usepackage{siunitx}
\usepackage{colortbl}
\usepackage[x11names,table]{xcolor}
\usepackage[dvipsnames, table]{xcolor}
\usepackage[most]{tcolorbox}
\usepackage{seqsplit}
\usepackage{tabularx,array}
\newcolumntype{Y}{>{\centering\arraybackslash}X}

\begin{document}
\title{Didactic knowledge or Clinical Cases? How Data Types Shape Medical Large Language Models}
\titlerunning{Didactic vs Clinical Data for Medical LLMs}

\author{
Yuzheng Fan\inst{1}$^{\star}$ \and
Haochun Wang\inst{1}$^{\star}$ \and
Sendong Zhao\inst{1}$^{\dagger}$ \and
Xiao Han\inst{1} \and
Ming Ma\inst{1} \and
Bing Qin\inst{1}
}

\authorrunning{Y. Fan et al.}

\institute{
Research Center for Social Computing and Interactive Robotics,\\
Harbin Institute of Technology, Harbin 150001, China\\
\email{\{hcwang,sdzhao\}@ir.hit.edu.cn}
}

\maketitle

\begingroup
\renewcommand{\thefootnote}{$\star$}
\footnotetext{Equal contribution.}
\renewcommand{\thefootnote}{$\dagger$}
\footnotetext{Corresponding author.}
\endgroup

\begin{abstract}
Medical large language models are commonly trained on mixtures of didactic data (e.g., textbooks) and clinical data (e.g., patient records), yet how these data types differentially shape model capabilities remains unclear. We address this issue with token-matched experiments that vary the didactic-to-clinical ratio and analyze how data composition affects performance, capability profiles, and error patterns across knowledge-intensive and clinic-oriented tasks. We uncover an asymmetric transfer across task types: clinical data improves clinic-oriented tasks while remaining competitive on knowledge-intensive ones, whereas didactic data mainly improves knowledge-intensive tasks. Error analysis suggests a knowing–doing gap, where improvements in knowledge recall do not reliably generalize to clinical reasoning. We further observe that modest amounts of clinical data yield most of the gains on EHR-grounded tasks, while the optimal mixture ratio varies with the knowledge and clinical reasoning demands of downstream tasks. These findings suggest that medical LLM data curation should be application-driven, with higher proportions of clinical data preferred for reasoning-intensive use cases.

\end{abstract}

\section{Introduction}

Medical large language models (LLMs) are trained on heterogeneous data sources---textbooks, guidelines, exam questions, and electronic health records (EHRs)---often pooled into unified corpora. This recipe has delivered strong benchmark performance \cite{singhal_2023}. However, it implicitly treats different training data sources as interchangeable. A fundamental question therefore remains underexplored: \textit{how do different types of medical data shape the capabilities of medical LLMs?}

\begin{figure}
  \includegraphics[width=\textwidth]{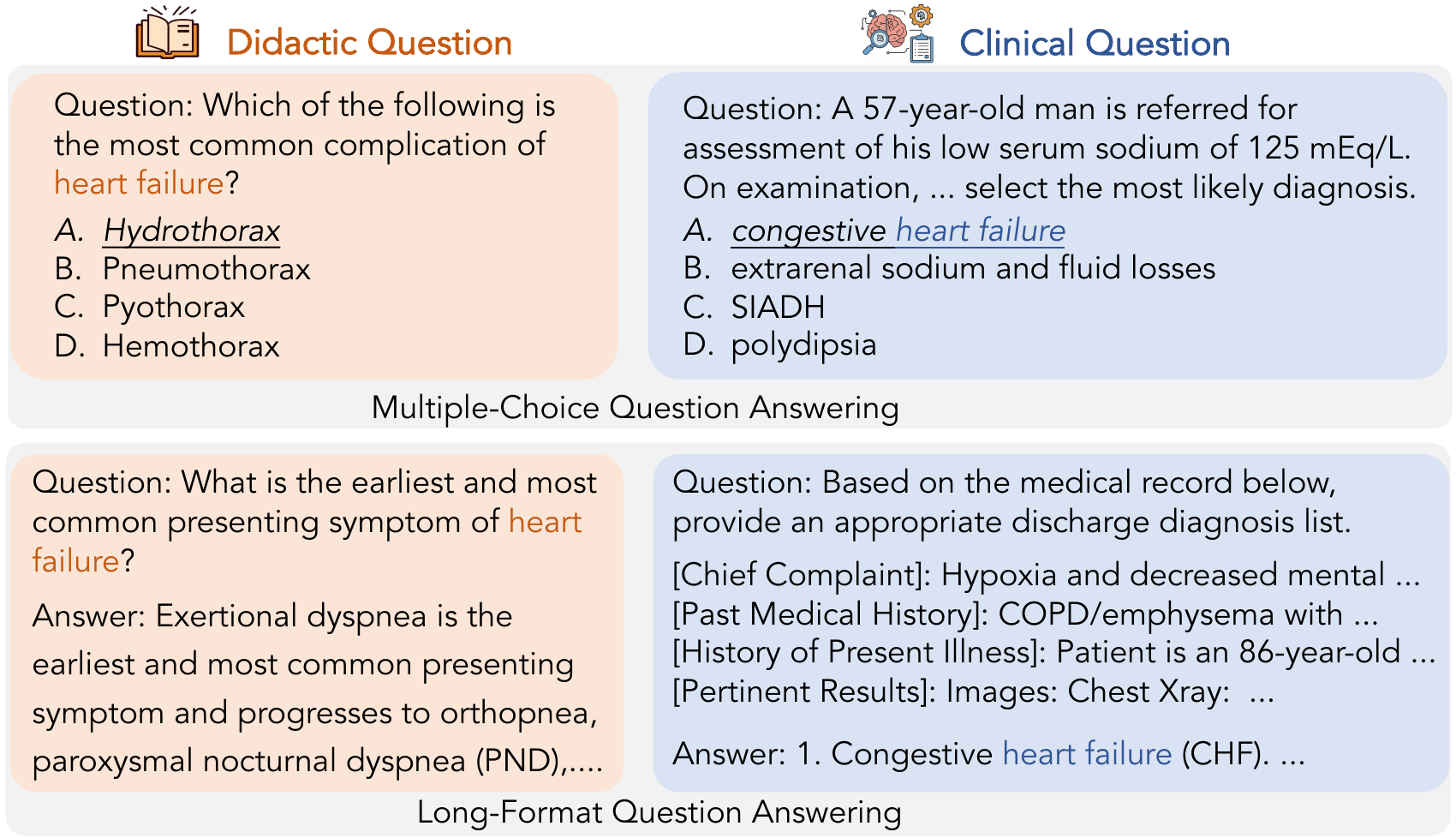}
  \caption{Examples of didactic and clinical case-based data across two QA formats.}
  \label{fig:example}
\end{figure}

Crucially, medical knowledge and practice evolve rapidly as new drugs, guidelines, and treatment protocols emerge, which motivates continual training and periodic updates to maintain currency in deployed systems. In this setting, understanding how different data types contribute to deployment-relevant capabilities is crucial. Data sources differ not only in surface form, but also in the information they provide and the reasoning they elicit. For example, a textbook passage on heart failure typically presents decontextualized mechanisms and definitions, whereas an ICU note about acute decompensation is a patient-grounded narrative where evidence unfolds over time and decisions are made under incomplete information. If such differences are systematic, then data composition may shape the capabilities of medical LLMs, leading to uneven gains across task types and deployment settings. To study this effect, we categorize medical corpora into two types as shown in Figure~\ref{fig:example}. We define \textit{didactic data} as decontextualized medical content without patient context, and \textit{clinical data} as patient-grounded scenarios in the training corpora of medical LLMs, inspired by the declarative and procedural knowledge in medical education \cite{Anderson_2013,Schmidt_2007}.

Prior work adapts LLMs to medicine via continual pretraining and fine-tuning on mixture of didactic and clinical corpora \cite{christophe_2024}. However, the contribution of each data type under fixed training budgets remains unclear. To address this, we train model variants under token-matched conditions, varying only the didactic-to-clinical ratio. We evaluate on benchmarks stratified into knowledge-intensive and clinic-oriented tasks. Clinic-oriented evaluation includes reasoning-intensive MCQs and ClinicalBench, a stagewise EHR-grounded suite construct from MIMIC-III \cite{MIMIC_2016} to assess models under realistic information constraints.

Our experiments reveal \textit{asymmetric transfer} across task types: increasing the share of clinical data improves performance on both clinic-oriented and knowledge-intensive tasks, whereas didactic data primarily benefits knowledge-intensive evaluation. This transfer is even more pronounced on EHR-grounded tasks, where didactic data yields limited gains. We also observe that a small amount of clinical data captures most of the improvements. To understand the mechanisms behind these patterns, we investigate three research questions (RQ):

\begin{itemize}
    \item \textbf{RQ1}: Do didactic and clinical data induce different reasoning directions?
    \item \textbf{RQ2}: How does data composition interact with the knowledge and reasoning demands of downstream tasks?
    \item \textbf{RQ3}: Do didactic and clinical data lead to different error patterns?
\end{itemize}

\vspace{0.3em}

Our contributions can be summarized as follows:

(1)~We introduce a token-matched experimental framework that isolates data composition as the causal variable, enabling controlled analysis of how different medical data corpora and mixture ratios shape the capabilities of medical LLMs.

(2)~We construct ClinicalBench, a suite of clinical tasks derived from EHRs with temporal masking, to evaluate clinical reasoning under realistic information constraints.

(3)~We provide fine-grained analysis revealing how data composition correlates with reasoning direction specialization and error distributions, offering mechanistic insights into the effects of different medical data types.
\section{Related Works}

\subsection{Medical LLMs}
Medical LLMs differ primarily in their base architectures and training data source. Didactic-centric models train on formalized knowledge like PubMed papers~\cite{BioGPT,BioMistral}, textbooks~\cite{PMC-LLAMA}, and structured medical knowledge bases~\cite{wang2025knowledge}, while clinical-centric models leverage real-world clinical narratives~\cite{GatorTron} and EHRs~\cite{clinicalGPT}. Recent efforts further incorporate clinical dialogues and multimodal data~\cite{MeLlama,Tu_2025,Llava-Meds}, with complementary approaches augmenting clinical reasoning through experience or memory-based mechanisms~\cite{han2026gsem}. While prior works generally mix both corpora, the specific impact of each source remains underexplored. We explicitly isolate these data types to examine their distinct effects on model capabilities.

\subsection{Evaluation Paradigms for Medical LLMs}

Medical LLMs are commonly evaluated on exam-style MCQ benchmarks testing knowledge recall~\cite{medqa,medbench,singhal_2023}. However, traditional n-gram metrics fail to capture clinical nuance, spurring rubric-based evaluation~\cite{abacha_2023,HealthBench}. EHR-grounded benchmarks~\cite{EHRNoteQA,emrQA} further extend evaluation to evidence-based QA and clinical text generation. Overall, the evaluation of medical LLMs is shifting from multiple-choice accuracy to multi-dimensional frameworks spanning QA suites, multi-turn dialogue, and EHR-based tasks, with increasing emphasis on safety and clinical utility, including the epistemic robustness of models whose correct beliefs may collapse under clinical pressure~\cite{xiao2026correct}.

\section{Study Design: Data Typology, Evaluation, and Analysis}

\begin{figure}
  \includegraphics[width=\textwidth]{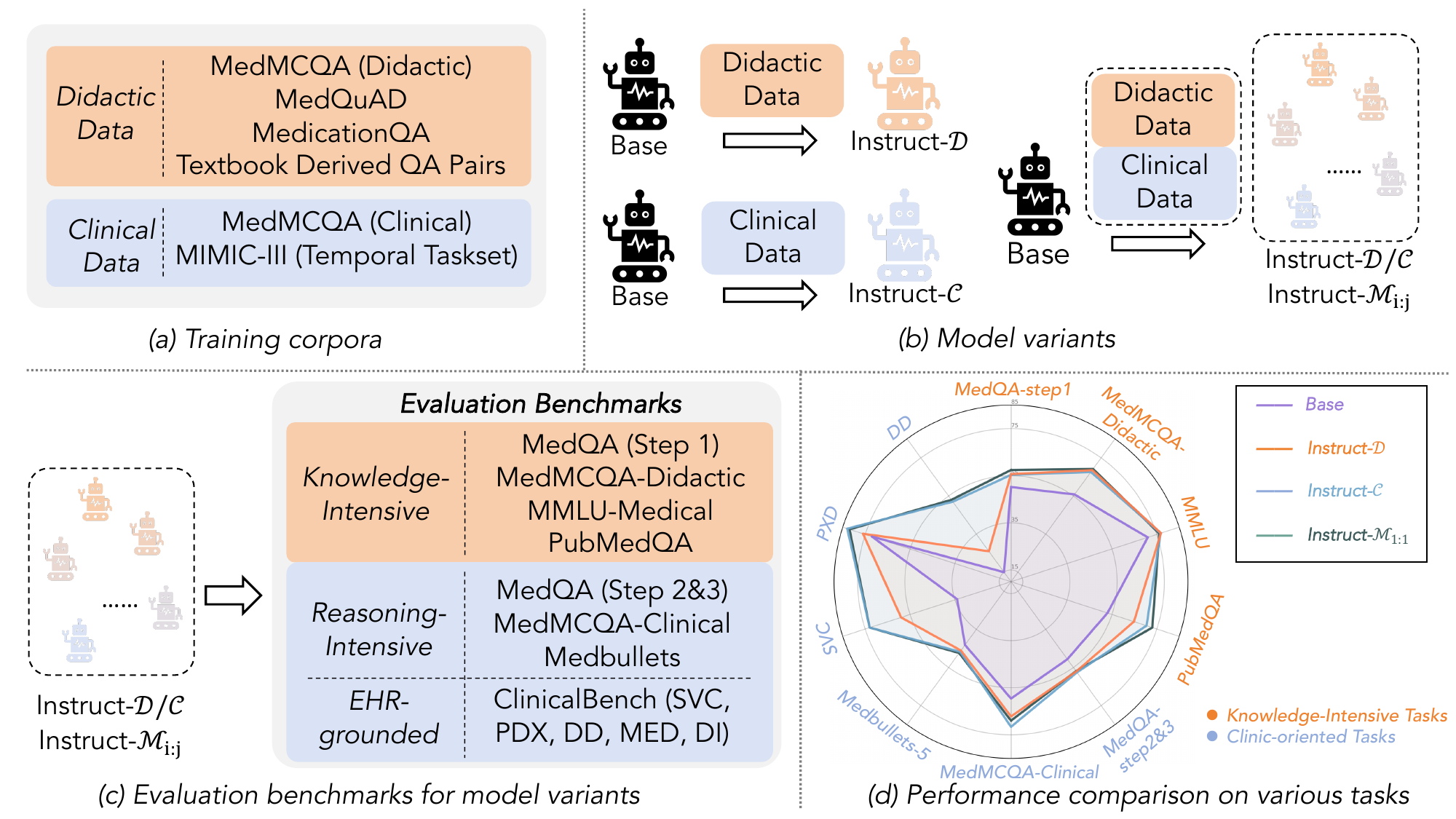}
  \caption{Overview of the study setup, including (a) the training corpora, (b) model variants fine-tuned on different data compositions, (c) evaluation benchmarks, and (d) comparative performance results for Qwen2.5-7B.}
  \label{fig:overall}
\end{figure}

In this section, we present a comparative framework to analyze how different data types influence medical LLM capabilities. We distinguish between didactic and clinical data, introduce a set of evaluation benchmarks, and describe analytical methods for deriving mechanistic insights into performance. Figure~\ref{fig:overall} provides an overview of this setup.

\subsection{Data Corpora}

We categorize medical corpora into two functional types, inspired by the distinction between declarative and procedural knowledge in medical education~\cite{Schmidt_2007}.

\paragraph{Didactic Data} This category comprises training examples that emphasize generalizable medical knowledge without patient-specific context, focusing on \textit{what is true} about medical concepts rather than \textit{how} to apply them in clinical practice. We assemble didactic data from four sources: MedMCQA (Didactic) \cite{MedMCQA}, MedQuAD \cite{MedQuAD}, MedicationQA, and explanatory passages and derived QA pairs from openly licensed medical textbooks.

\paragraph{Clinical Data} Clinical data embeds medical knowledge within clinical narratives characterized by temporal progression, incomplete information, diagnostic uncertainty, and the practical constraints of healthcare delivery. We construct clinical data from two sources. (1)~MedMCQA (Clinical): vignette-containing questions from MedMCQA presenting patient demographics, symptoms, or examination findings followed by diagnostic or management queries. These questions provide examples of systematic clinical reasoning. (2)~MIMIC-III with Temporal Care Simulation: We apply a temporal masking strategy to MIMIC-III EHRs to simulate real-world clinical dynamics. At each care stage (admission, hospitalization, discharge), the model accesses only information available at that point, forcing medical LLMs to reason under uncertainty—the central challenge of clinical decision-making. We design seven task types covering diagnosis, treatment, and disposition prediction.

\subsection{Medical LLM Capabilities}

We characterize medical LLM capabilities along two primary dimensions. \textit{Knowledge Recall} is the ability to retrieve and correctly apply canonical, decontextualized medical knowledge, including definitions, diagnostic criteria ,and pathophysiological mechanisms. \textit{Clinical Reasoning} is the ability to integrate patient-specific information and reason under uncertainty, including synthesizing findings across a narrative, tracking temporal evolution, and producing context-appropriate decisions.

\subsection{Evaluation for Model Variants}

We organize benchmarks by capability demand into knowledge-intensive and clinic-oriented tasks. Clinic-oriented evaluation covers both MCQ-based and EHR-grounded settings.

\paragraph{Knowledge-Intensive Benchmarks} MCQ benchmarks test knowledge recall and conceptual understanding ability, including MedQA Step~1 \cite{medqa}, MedMCQA-Didactic, MMLU-Medical \cite{mmlu}, and PubMedQA \cite{pubmedqa}.

\paragraph{Reasoning-Intensive Benchmarks} These benchmarks require clinical reasoning and decision-making ability over patient vignettes, including MedQA Step~2\&3, MedMCQA-Clinical, and Medbullets \cite{medbullets}.

\paragraph{ClinicalBench (EHR-Grounded)} We introduce ClinicalBench, constructed from MIMIC-III under the temporal masking framework, where models generate clinical decisions from patient EHRs under realistic information constraints. It comprises five tasks: \textbf{Service Prediction} (\textsc{Svc}), \textbf{Primary Diagnosis} (\textsc{Pdx}), \textbf{Discharge Disposition} (\textsc{Dd}), \textbf{Discharge Medication} (\textsc{Med}), and \textbf{Discharge Instruction} (\textsc{Di}).

\begin{table}
\caption{Performance on knowledge-intensive and reasoning-intensive benchmarks across model variants. \textbf{Bold} indicates the best performance.}
\label{tab:MCQs}
\centering
\resizebox{\textwidth}{!}{%
\begin{tabular}{|l|l|c|c|c|c|c|c|c|c|}
\hline
\multirow{2}{*}{Base Model} &
\multirow{2}{*}{Model Variant} &
\multicolumn{3}{c|}{MedQA} &
\multicolumn{2}{c|}{MedMCQA} &
PubMedQA & MMLU & Medbullets-5 \\
\cline{3-7}
& & step1 & step2\&3 & All & Didactic & Clinical & & & \\
\hline
\multirow{6}{*}{Qwen2.5-7B}
  & Base                         & 50.22 & 50.67 & 50.43 & 55.80 & 59.60 & 53.00 & 70.98 & 43.18 \\
  & Instruct-$\mathcal{D}$       & 55.67 & 57.07 & 56.32 & 68.60 & 67.20 & 64.80 & 76.85 & 46.10 \\
  & Instruct-$\mathcal{C}$       & 55.38 & \textbf{57.58} & 56.40 & 67.60 & \textbf{71.60} & 70.40 & 76.68 & 46.75 \\
  & Instruct-$\mathcal{M}_{1:3}$ & 55.96 & 57.07 & 56.48 & 66.80 & 70.00 & 72.80 & 76.21 & 48.70 \\
  & Instruct-$\mathcal{M}_{1:1}$ & \textbf{57.44} & 56.90 & \textbf{57.19} & \textbf{69.20} & 69.00 & 73.00 & 76.12 & 47.40 \\
  & Instruct-$\mathcal{M}_{3:1}$ & 55.82 & 56.90 & 56.32 & 68.80 & 68.80 & \textbf{74.40} & \textbf{77.13} & \textbf{49.03} \\
\hline
\multirow{6}{*}{Llama3.1-8B}
  & Base                         & 52.43 & 55.72 & 53.97 & 69.40 & 68.00 & 75.00 & 70.71 & 46.43 \\
  & Instruct-$\mathcal{D}$       & 51.55 & 55.39 & 53.34 & 73.00 & 69.20 & 74.60 & \textbf{72.64} & 45.78 \\
  & Instruct-$\mathcal{C}$       & 51.10 & \textbf{58.42} & 54.52 & 72.60 & \textbf{72.20} & 71.80 & 71.07 & 44.16 \\
  & Instruct-$\mathcal{M}_{1:3}$ & 53.46 & 54.38 & 53.89 & \textbf{74.60} & 70.60 & \textbf{75.20} & 71.44 & 47.08 \\
  & Instruct-$\mathcal{M}_{1:1}$ & 52.28 & 56.23 & 54.12 & 71.60 & 71.60 & 74.80 & 71.81 & 45.78 \\
  & Instruct-$\mathcal{M}_{3:1}$ & \textbf{54.20} & 56.40 & \textbf{55.22} & 73.60 & 71.80 & 74.00 & 71.26 & \textbf{47.40} \\
\hline
\end{tabular}
}
\end{table}

\begin{table}
\caption{Performance on ClinicalBench across model variants. $^{*}$ indicates official instruct model. \textbf{Bold} indicates the best performance per task within each base model.}
\label{tab:clinicalbench}
\centering
\resizebox{\textwidth}{!}{%
\begin{tabular}{|l|l|c|c|c|c|c|c|}
\hline
\multirow{2}{*}{Base Model} &
\multirow{2}{*}{Model Variant} &
\multirow{2}{*}{Service} &
\multirow{2}{*}{\shortstack{Primary\\Diagnosis}} &
\multicolumn{2}{c|}{Medication} &
\multirow{2}{*}{\shortstack{Discharge\\Disposition}} &
\multirow{2}{*}{\shortstack{Discharge\\Instruction}} \\
\cline{5-6}
& & & & F1 & Jaccard & & \\
\hline

\multirow{7}{*}{Qwen2.5-7B}
  & Base                          & 34.00 & 72.00 & 19.37 & 10.72 & 15.00 & 2.14 \\
  & Instruct$^{*}$                & 50.00 & 75.00 & 16.01 &  8.70 & 43.00 & 2.51 \\
  & Instruct-$\mathcal{D}$        & 59.00 & 76.00 & 16.15 &  8.78 & 26.00 & 2.08 \\
  & Instruct-$\mathcal{C}$        & 73.00 & \textbf{83.00} & \textbf{30.16} & \textbf{17.76} & 52.00 & 2.53 \\
  & Instruct-$\mathcal{M}_{1:3}$  & \textbf{78.00} & 80.00 & 26.93 & 15.56 & 49.00 & 2.49 \\
  & Instruct-$\mathcal{M}_{1:1}$  & 73.00 & 82.00 & 28.18 & 16.40 & \textbf{53.00} & \textbf{2.57} \\
  & Instruct-$\mathcal{M}_{3:1}$  & 70.00 & \textbf{83.00} & 23.02 & 13.01 & \textbf{53.00} & 2.30 \\
\hline

\multirow{7}{*}{Llama3.1-8B}
  & Base                          & 68.00 & 56.00 & 14.25 &  7.67 & 16.00 & 1.54 \\
  & Instruct$^{*}$                & 61.00 & 62.00 & 15.04 &  8.13 & 44.00 & 2.58 \\
  & Instruct-$\mathcal{D}$        & 77.00 & 55.00 & 17.87 &  9.81 & 31.00 & 1.87 \\
  & Instruct-$\mathcal{C}$        & \textbf{81.00} & 72.00 & 21.43 & 12.00 & \textbf{57.00} & 2.45 \\
  & Instruct-$\mathcal{M}_{1:3}$  & 80.00 & \textbf{76.00} & 19.75 & 10.96 & 40.00 & \textbf{2.79} \\
  & Instruct-$\mathcal{M}_{1:1}$  & 78.00 & 60.00 & \textbf{22.57} & \textbf{12.72} & 52.00 & 2.48 \\
  & Instruct-$\mathcal{M}_{3:1}$  & 72.00 & 73.00 & 19.43 & 10.76 & 46.00 & 2.14 \\
\hline
\end{tabular}
}
\end{table}

\subsection{Analysis Beyond Performance}

Beyond aggregate benchmark scores, we conduct three complementary analyses to investigate the mechanisms underlying performance differences between didactic and clinical training corpora: (i)~probing directional reasoning preferences (RQ1), (ii)~examining how task demands interact with data composition (RQ2), and (iii)~characterizing failure patterns (RQ3).

\paragraph{Directional Reasoning} Didactic data encourages \textit{Concept$\rightarrow$Attributes} reasoning (mapping concepts to observable properties), while clinical data emphasizes \textit{Evidence$\rightarrow$Diagnosis} inference (from findings to conclusions). We operationalize each direction with two probing tasks: Term$\rightarrow$Symptoms and Term$\rightarrow$ Definition for the former; Symptoms$\rightarrow$Diagnosis and EHR$\rightarrow$Diagnosis (\textsc{Pdx}) for the latter. To quantify directional preference, we define the Direction Bias Index (DBI):
\begin{equation}
    \text{DBI} = \frac{\bar{S}_{\text{C} \rightarrow \text{A}} - \bar{S}_{\text{E} \rightarrow \text{D}}}{\bar{S}_{\text{all}}}
\end{equation}
where $\bar{S}_{\text{C}\rightarrow\text{A}}$ and $\bar{S}_{\text{E}\rightarrow\text{D}}$ are mean scores for each direction and $\bar{S}_{\text{all}}$ is the overall mean. Positive DBI indicates concept-oriented bias; negative indicates evidence-to-diagnosis preference.

\paragraph{Demand Decomposition} Aggregate benchmark scores quantify overall performance, but they do not reveal which demand profiles are most sensitive to corpus composition. To localize these effects, we decompose each MCQ item along two axes: knowledge demand ($K$) and reasoning demand ($R$). Then, we partition the evaluation set into four quadrants ($K^\pm R^\pm$). We report accuracy for each quadrant and assess (i) which data composition results in the largest improvements within each quadrant, and (ii) whether mixed corpora exhibit complementarity beyond a ratio-weighted linear expectation.

\paragraph{Failure Analysis} Accuracy indicates whether a model answers correctly, but it does not explain what types of error are made. To test whether corpus composition is associated with systematic differences in failure patterns, we annotate each incorrectly answered question with a single primary error cause. We use four categories: knowledge errors ($E_{\text{Know.}}$), reasoning errors ($E_{\text{Reas.}}$), context errors ($E_{\text{Ctx.}}$), and format errors ($E_{\text{Fmt.}}$).

\section{Experiments}

\subsection{Experimental Setup}

\paragraph{Base Models} We conduct experiments on Qwen2.5-7B-Base and Llama3.1-8B-Base. We choose base models rather than instruction-tuned variants to ensure a clean experimental setup. All observed capabilities can be attributed to our controlled supervision.

\paragraph{Training Variants} We train five variants for each base model: didactic-only ($\mathcal{D}$), clinical-only ($\mathcal{C}$), and three mixtures at ratios 1:3, 1:1, and 3:1 ($\mathcal{M}_{1:3}$, $\mathcal{M}_{1:1}$, $\mathcal{M}_{3:1}$), all under the same token budget, schedule, and hyperparameters.

\paragraph{Evaluation Metrics} For MCQs and ClinicalBench classification tasks (\textsc{Svc}, \textsc{Pdx}, \textsc{Dd}), we report accuracy. For task \textsc{Med}, we use Jaccard and micro-F1; for task \textsc{Di}, we adopt multi-dimensional GPT-based evaluation~\cite{CMBbench} assessing accuracy, clarity, faithfulness, and completeness.

\subsection{Overall Performance}
\label{sec:overall}

Tables~\ref{tab:MCQs} and~\ref{tab:clinicalbench} present results across medical QA benchmarks and ClinicalBench. The two corpora yield asymmetric transfer across evaluation settings. On knowledge-intensive QA, Instruct-$\mathcal{D}$ holds a slight advantage on some benchmarks (e.g., MMLU-Medical), while Instruct-$\mathcal{C}$ performs better on reasoning-intensive MCQs and dominates across all ClinicalBench tasks. Instruct-$\mathcal{D}$ yields limited and unstable gains on EHR-grounded settings, whereas Instruct-$\mathcal{C}$ remains competitive on knowledge-intensive QA while excelling in EHR contexts. This reveals an \emph{asymmetric transfer}: clinical data supports both task types, while didactic data primarily benefits knowledge-intensive evaluation—suggesting clinical data reinforces factual knowledge through application, whereas didactic data does not cultivate the procedural competencies required for EHR-grounded reasoning. Mixed corpora provide complementary benefits, with mixture variants outperforming both single-source counterparts on several benchmarks. On ClinicalBench, a small clinical fraction captures most of the improvement over Instruct-$\mathcal{D}$, with diminishing returns from further increases. The optimal ratio is task-dependent: Instruct-$\mathcal{M}_{1:3}$ favors EHR-grounded tasks, while Instruct-$\mathcal{M}_{3:1}$ is competitive on knowledge-intensive QA—suggesting corpus composition should be selected based on the target use case.

\subsection{Directional Reasoning Analysis (RQ1)}

To avoid confounding data-type effects with coverage differences, we restrict this analysis to diseases appearing in both corpora. We evaluate Symptoms$\rightarrow$Diagnosis on both typical and atypical symptom sets, using the atypical subset to probe robustness to clinical presentations. As shown in Table~\ref{tab:rq1}, results confirm our hypothesis: Instruct-$\mathcal{D}$ outperforms Instruct-$\mathcal{C}$ on \textit{Concept$\rightarrow$Attributes} tasks, while Instruct-$\mathcal{C}$ excels on \textit{Evidence$\rightarrow$Diagnosis} tasks across both typical and atypical presentations. The DBI captures this shift: starting from a near-balanced base (DBI~$\approx 0$), Instruct-$\mathcal{D}$ yields a positive DBI and Instruct-$\mathcal{C}$ a negative DBI, while Instruct-$\mathcal{M}_{1:1}$ remains near zero. This directional specialization explains the asymmetric transfer in Section~\ref{sec:overall}: clinical data implicitly exercises concept knowledge through diagnostic application, enabling transfer to knowledge-intensive tasks; didactic data lacking such inferential practice does not develop the synthesis capabilities required for clinical reasoning.

\begin{table}
\caption{Performance on each directional mapping task on Qwen2.5-7B. A positive DBI indicates concept-oriented bias; a negative indicates evidence-to-diagnosis preference. Sym., Def., and Diag. stand for Symptoms, Definition, and Diagnosis respectively. \textbf{Bold} indicates the best performance across model variants.}
\label{tab:rq1}
\centering
\begin{tabular}{|l|c|c|c|c|c|c|}
\hline
\multirow{2}{*}{Model Variant} &
\multirow{2}{*}{Term$\rightarrow$Sym.} &
\multirow{2}{*}{Term$\rightarrow$Def.} &
\multicolumn{2}{c|}{Sym.$\rightarrow$Diag.} &
\multirow{2}{*}{EHR$\rightarrow$Diag.} &
\multirow{2}{*}{DBI} \\
\cline{4-5}
& & & Typical & Atypical & & \\
\hline
Base                          & 53.18 & 41.98 & 26.83 & 17.56 & 72.00 & {\color{ForestGreen!40!gray}+0.01} \\
\hline
Instruct-$\mathcal{D}$        & \textbf{64.70} & \textbf{48.40} & 30.24 & 24.39 & 76.00 & {\color{ForestGreen!90!black}+0.09} \\
Instruct-$\mathcal{C}$        & 57.72 & 47.00 & \textbf{35.12} & \textbf{28.29} & \textbf{83.00} & {\color{red!90!black}-0.09} \\
Instruct-$\mathcal{M}_{1:1}$  & 62.80 & 46.72 & 34.63 & 24.39 & 82.00 & {\color{red!50!gray}-0.02} \\
\hline
\end{tabular}
\end{table}

\subsection{Quadrant-Wise Analysis (RQ2)}

\paragraph{Quadrant-Wise Performance} Table~\ref{tab:quadrant} suggests that different quadrants favor different data compositions. In the reasoning-dominant quadrant ($K^-R^+$), Instruct-$\mathcal{M}_{1:3}$ achieves the highest accuracy, which is consistent with clinical data being more helpful when inference is required despite limited explicit knowledge in the prompt. In the knowledge-dominant quadrant ($K^+R^-$), Instruct-$\mathcal{M}_{1:1}$ performs best, indicating that combining the two corpora can be beneficial even when the question format is not strongly reasoning-intensive. In the dual-demand quadrant ($K^+R^+$), Instruct-$\mathcal{C}$ yields the strongest result, suggesting that evidence-driven inference can be a limiting factor when both knowledge and reasoning requirements are high. Finally, in the low-demand quadrant ($K^-R^-$), Instruct-$\mathcal{M}_{3:1}$ performs best, indicating that when questions require limited multi-step inference, additional didactic exposure is sufficient and may even be advantageous.

\begin{table}
\caption{Performance across knowledge-reasoning quadrants on Qwen2.5-7B. \textbf{Bold} indicates the best performance per quadrant.}
\label{tab:quadrant}
\centering
\begin{tabular}{|l|c|c|c|c|}
\hline
\textbf{Model Variant} &
\textbf{$K^-R^-$} &
\textbf{$K^-R^+$} &
\textbf{$K^+R^-$} &
\textbf{$K^+R^+$} \\
\hline
Base                          & 72.50 & 73.01 & 58.10 & 44.00 \\
Instruct-$\mathcal{D}$        & 83.85 & 73.01 & 69.83 & 48.33 \\
Instruct-$\mathcal{C}$        & 82.13 & 74.60 & 70.94 & \textbf{53.00} \\
Instruct-$\mathcal{M}_{1:3}$  & 82.13 & \textbf{76.19} & 67.60 & 50.33 \\
Instruct-$\mathcal{M}_{1:1}$  & 82.82 & 74.60 & \textbf{73.18} & 51.66 \\
Instruct-$\mathcal{M}_{3:1}$  & \textbf{84.87} & 73.01 & 71.50 & 50.00 \\
\hline
\end{tabular}
\end{table}

\paragraph{Synergy Analysis} We define the Synergy Index (SI) as the deviation of observed mixture performance from a ratio-weighted linear interpolation of the two single-source variants:

\begin{equation}
    \text{SI} = \text{Acc}_{\mathcal{M}} - \left( \alpha \cdot \text{Acc}_{\mathcal{D}} + (1-\alpha) \cdot \text{Acc}_{\mathcal{C}} \right)
\end{equation}
where $\alpha$ is the didactic proportion. As shown in Table~\ref{tab:synergy}, Instruct-$\mathcal{M}_{1:3}$ exceeds the linear expectation in $K^-R^+$ but falls below it in $K^+R^-$, with Instruct-$\mathcal{M}_{3:1}$ showing the opposite pattern. These results indicate that mixture optimization is not about finding a universally superior ratio, but about matching composition to task demands: misaligned mixtures underperform linear expectations, while well-matched ones unlock complementary benefits.

\begin{table}
\caption{Synergy analysis with ratio-weighted linear expectation. Positive values indicate performance exceeding the mixture-proportion linear baseline.}
\label{tab:synergy}
\centering
\begin{tabular}{|l|c|c|c|c|}
\hline
\textbf{Model Variant} & \textbf{$K^- R^-$} & \textbf{$K^- R^+$} & \textbf{$K^+ R^-$} & \textbf{$K^+ R^+$} \\
\hline
Instruct-$\mathcal{M}_{1:3}$ & {\color{red!80!black}-0.43} & {\color{teal}+1.99} & {\color{red!80!black}-3.06} & {\color{red!80!black}-1.50} \\
Instruct-$\mathcal{M}_{1:1}$ & {\color{red!80!black}-0.17} & {\color{teal}+0.80} & {\color{teal}+2.80} & {\color{teal}+1.00} \\
Instruct-$\mathcal{M}_{3:1}$ & {\color{teal}+1.45} & {\color{red!80!black}-0.40} & {\color{teal}+1.39} & {\color{teal}+0.50} \\
\hline
\end{tabular}
\end{table}

\subsection{Failure Mode Analysis (RQ3)}
\label{sec:rq3}

\begin{figure}
  \centering
  \includegraphics[width=\textwidth]{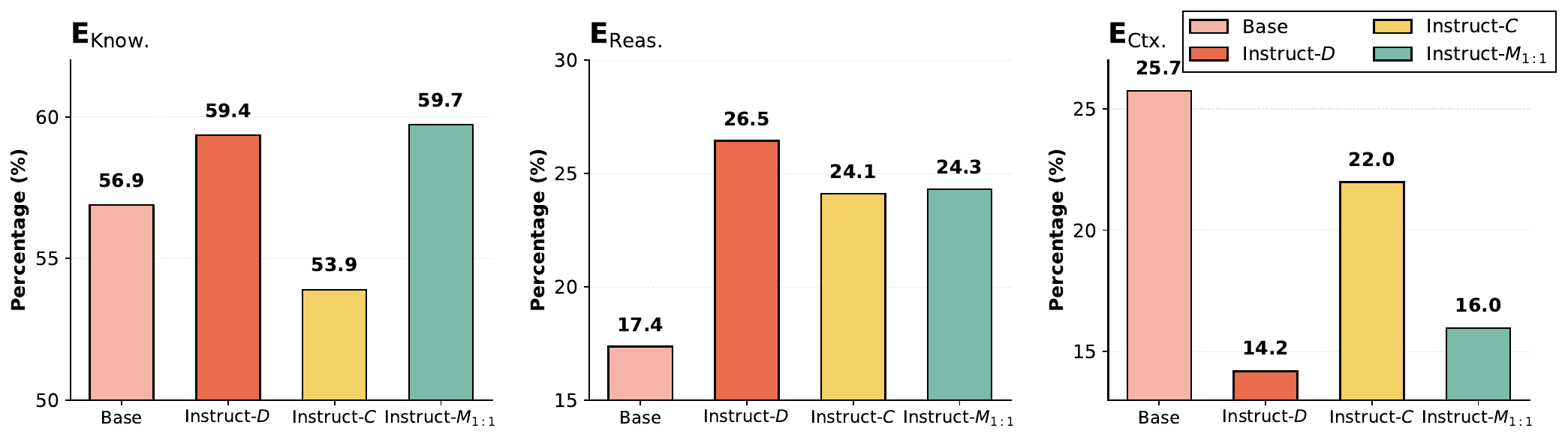}
  \caption{Error distribution in the $K^+R^+$ quadrant. Bars show the proportion of each error type within the total errors for each variant in this quadrant.}
  \label{fig:error-kprp}
\end{figure}

Table~\ref{tab:error-dist} shows that knowledge errors constitute the majority of failures across all variants, but error profiles differ by data composition. Instruct-$\mathcal{D}$ shows a higher share of reasoning errors and lower share of context errors than the base model, suggesting a \textit{knowing-doing gap}—knowledge is acquired but not effectively applied. Instruct-$\mathcal{C}$ reduces both total errors and the share of knowledge errors, reinforcing that clinical contexts strengthen knowledge recall through application. Instruct-$\mathcal{M}_{1:1}$ falls between the two. This pattern is most pronounced in the $K^+R^+$ quadrant (Figure~\ref{fig:error-kprp}), where Instruct-$\mathcal{D}$ yields the highest reasoning error rate and Instruct-$\mathcal{C}$ the lowest knowledge error rate. Overall, the results suggest that clinical data improves robustness on errors tied to evidence-driven inference, while didactic data more often struggles when questions require synthesizing multiple cues, even when factual knowledge is available.

\begin{table}
\caption{Error type distribution across data compositions, where $E_{\text{Know.}}$, $E_{\text{Reas.}}$, $E_{\text{Ctx.}}$, and $E_{\text{Fmt.}}$ denote knowledge, reasoning, context, and format errors, respectively. Percentages indicate the proportion within total errors for each model.}
\label{tab:error-dist}
\centering
\begin{tabular}{|l|c|c|c|c|}
\hline
\textbf{Model Variant} & $E_{\text{Know.}}$ & $E_{\text{Reas.}}$ & $E_{\text{Ctx.}}$ & $E_{\text{Fmt.}}$ \\
\hline
Base
  & 229 {\scriptsize(67.4\%)} & 38 {\scriptsize(11.2\%)} & 60 {\scriptsize(17.6\%)} & 13 {\scriptsize(3.8\%)} \\
Instruct-$\mathcal{D}$
  & 195 {\scriptsize(71.4\%)} & 51 {\scriptsize(18.7\%)} & 24 {\scriptsize(8.8\%)}  &  3 {\scriptsize(1.1\%)} \\
Instruct-$\mathcal{C}$
  & 182 {\scriptsize(69.7\%)} & 38 {\scriptsize(14.6\%)} & 34 {\scriptsize(13.0\%)} &  7 {\scriptsize(2.7\%)} \\
Instruct-$\mathcal{M}_{1:1}$
  & 185 {\scriptsize(70.6\%)} & 43 {\scriptsize(16.4\%)} & 30 {\scriptsize(11.5\%)} &  4 {\scriptsize(1.5\%)} \\
\hline
\end{tabular}
\end{table}
\section{Conclusion}

In this study, we examine how didactic and clinical corpora shape medical LLM capabilities under token-matched experiments. Clinical data performs strongly on clinic-oriented tasks while remaining competitive on knowledge-intensive ones, whereas didactic data yields limited gains on clinic-oriented evaluation—revealing an asymmetric transfer consistent with a \textit{knowing--doing gap}. The optimal composition varies with task demands and a modest clinical fraction captures most gains on EHR-grounded tasks. These findings highlight the importance of composition-aware, application-driven data curation for medical LLMs.

\begin{credits}
\subsubsection{\ackname}
This work was supported in part by the National Natural Science Foundation of China (Grant No. 62576126) and the Key R\&D Program of Heilongjiang Province (Grant No. 2023ZX01A11).
\end{credits}

\bibliographystyle{splncs04}
\bibliography{custom}

\end{document}